\documentclass[a4paper]{article}
\usepackage{algorithm,algpseudocode,amsmath,url} 
\usepackage[]{gnuplottex}
\usepackage{graphicx,fullpage,placeins}
\graphicspath{{.}{figures/}}
\usepackage[sort&compress,authoryear,semicolon,square]{natbib}
\renewcommand{\cite}{\citep}
\newtheorem{definition}{Definition}

\usepackage{silence}
\usepackage[font=footnotesize, labelfont=sf]{caption}
\usepackage[font=footnotesize, labelfont=sf]{subcaption}
\begin{document}  

\title{The Predictive Context Tree: \\Predicting Contexts and Interactions}
\author{Alasdair Thomason, Nathan Griffiths, Victor Sanchez\\
~\\
Department of Computer Science, \\University of Warwick, UK}
\date{September 2016} 

\maketitle

\let\svthefootnote\thefootnote
\let\thefootnote\relax\footnote{Authors' contact details: \{Alasdair.Thomason, Nathan.Griffiths, V.F.Sanchez-Sliva\}@warwick.ac.uk\\}
\addtocounter{footnote}{-1}\let\thefootnote\svthefootnote

\begin{abstract}
With a large proportion of people carrying location-aware smartphones, we have an unprecedented platform from which to understand individuals and predict their future actions. This work builds upon the \emph{Context Tree} data structure~\cite{Thomason:ContextTrees} that summarises the historical contexts of individuals from augmented geospatial trajectories, and constructs a predictive model for their likely future contexts. The \emph{Predictive Context Tree (PCT)} is constructed as a hierarchical classifier, capable of predicting both the future locations that a user will visit and the contexts that a user will be immersed within. The PCT is evaluated over real-world geospatial trajectories, and compared against existing location extraction and prediction techniques, as well as a proposed hybrid approach that uses identified land usage elements in combination with machine learning to predict future interactions. Our results demonstrate that higher predictive accuracies can be achieved using this hybrid approach over traditional extracted location datasets, and the PCT itself matches the performance of the hybrid approach at predicting future interactions, while adding utility in the form of context predictions. Such a prediction system is capable of understanding not only where a user will visit, but also their context, in terms of what they are likely to be doing.
\end{abstract} 

\maketitle
 


\section{Introduction}\label{sec:introduction}

Much existing work has focused on identifying locations from geospatial trajectories as a basis for prediction, aiming to determine the likely regions that an individual or other entity will visit in the future. While this is a useful component of many services, the identified locations do not necessarily correspond to real-world entities, often spanning multiple buildings or areas. Other avenues of research have focused on identifying activities or contexts that an individual has been immersed in, but these typically require data from low-level sensors (e.g.\ accelerometer and heart-rate) or video footage, neither of which are commonly available. Aiming to overcome both of these limitations, this work focuses on predicting both the future location and context of an individual, using only geospatial trajectories collected from or about an individual, in addition to data that can be added after the time of collection. The methods presented in this paper are building blocks for the creation of intelligent and tailored services.

Using the \emph{Context Tree}~\cite{Thomason:ContextTrees} data structure as a basis, this paper first discusses substituting existing location extraction techniques with element identification for prediction applications. Elements represent real-world locations and so provide additional insight and information that existing techniques do not consider, such as tags that describe each element's function. We then present the \emph{Predictive Context Tree (PCT)}, a new hierarchical classification model for predicting future element interactions and the future contexts that the user will be immersed within. An earlier version of the PCT was presented in~\cite{Thomason:ContextTreePrediction}, however this paper provides a greater level of detail, an expanded evaluation and extends the PCT to predict multiple land usage elements and contexts. We demonstrate the utility of the PCT through results showing increased predictive accuracies over existing techniques.

We begin with a discussion of related work in location and context identification and prediction in Section~\ref{sec:related}. Section~\ref{sec:augmentation} presents a modification to the context tree generation procedure (discussed in Section~\ref{sec:related:contexttree}) that allows for the identification of individual elements that represent real-world locations that a user interacted with, and proposes using these locations in lieu of extracted locations as a basis for location prediction using existing machine learning techniques. Section~\ref{sec:pct} then goes on to present a new technique for location prediction, the \emph{Predictive Context Tree} (PCT), that uses the information from clustered context trees to identify which identified location the user will visit in the future. The PCT is designed to additionally operate as a context predictor, where contexts instead of locations can be returned as prediction results. A quantitative investigation of predictions made over extracted locations and real-world elements, in addition to those predicted with existing techniques and the PCT is then conducted, with the methodology presented in Section~\ref{sec:methodology} and results in Section~\ref{sec:results}. Finally, the paper is concluded in Section~\ref{sec:conclusion}.


\section{Related Work}\label{sec:related} 

This section introduces existing methods for extracting locations from geospatial trajectories, and then considers predicting future locations and contexts. Finally, the \emph{Context Tree} data structure and generation framework is discussed, as this provides a foundation for the work in this paper.

\subsection{Location Extraction}

Locations are typically extracted from geospatial trajectories using two distinct clustering steps. The first step is performed by iterating over the trajectory to identify periods of low mobility. This is followed by a clustering step that groups the extracted \emph{visits} (or \emph{stops}) into \emph{locations}~\cite{Montoliu:2010bh,Zheng:2015fy}.

Extracting periods of low mobility, referred to as \emph{visit extraction}, was initially performed by looking for periods of missing data, which were assumed to correlate to a person being indoors, since early data logging devices did not function in enclosed spaces~\cite{Ashbrook:2003bi}. Modern devices, however, can use a combination of techniques to determine their location and are much more resilient. Taking account of this new hardware, time and distance thresholds can be applied to determine the maximum size of visits, where a visit must be longer than a specified duration and contained within a specified radius~\cite{Andrienko:2013cd,Kang:2004tf,Montoliu:2010bh,Toyama:2005wx,Zheng:2010cj,Zhou:2014eg}. However, these methods still suffer from a lack of resilience to noise, and so more specialised algorithms have been proposed to analyse the trend of motion of an individual, and determine when they are moving away from a visit location. Such techniques include the Gradient-based Visit Extractor (GVE)~\cite{Thomason:2016gv}. 

Once they have been extracted, visits are clustered together using standard approaches, including density-based techniques such as DBSCAN~\cite{Andrienko:2013cd,Ester:1996tm,Montoliu:2010bh} and partitional approaches, such as k-means~\cite{Ashbrook:2003bi,MacQueen:1967uv}. While both are applicable, DBSCAN is more commonly used because it does not require the number of locations to be known a priori and allows for arbitrary shaped and sized clusters.

\subsection{Location Prediction}

Making predictions over extracted locations allows for the provision of services that consider where a person will likely be in the future. Although location prediction was originally considered for preparing handovers in cellular telephony networks~\cite{Akoush:2007ct,Bilurkar:2002cf}, recent work has focused on predicting specific locations that will be visited by the user from sets of locations extracted from trajectories~\cite{Ashbrook:2003bi,Assam:2013di,Chon:2012kr,Toyama:2005wx}. Location prediction has been realised using support vector machines (SVMs)~\cite{Wang:2012ta}, neural networks~\cite{Akoush:2007ct,Bilurkar:2002cf,Thomason:2015po}, and Markov models~\cite{Ashbrook:2003bi,Assam:2013di,Toyama:2005wx}.

\subsection{Context Prediction}

While locations provide a basis for understanding users, identifying the context of user actions vastly increases the available information for making decisions about users. The aim of context identification is to detect periods of time in which the user likely had similar goals, or were performing the same task. Existing work has considered context extraction from sequence- and entropy-based approaches~\cite{Bao:2011hy,Lemlouma:2004cm}, used as a basis for adapting computing applications to a current context~\cite{Anagnostopoulos:2006ge,Lemlouma:2004cm}. 

Predicting the future context of an individual has also been considered, where location and context predictions are sometimes treated together. Le \emph{et al.}~\citeyear{Le:2015vv} use historical contexts to suggest locations a user may wish to visit, while Assam \emph{et al.}~\citeyear{Assam:2014cd} aim to use interaction histories to predict the types of contexts associated with each location. Aiming to solve the problem of location prediction, but using additional knowledge offered by context extraction, Bhyri \emph{et al.}~\citeyear{Bhyri:2015cz} propose a two-step approach to location prediction that first aims to predict the context that a user will be in and then aims to match this context to the most likely location the user will visit to fulfil the context, achieved through classification techniques.

\subsection{The Context Tree}\label{sec:related:contexttree}

\begin{figure}[t]
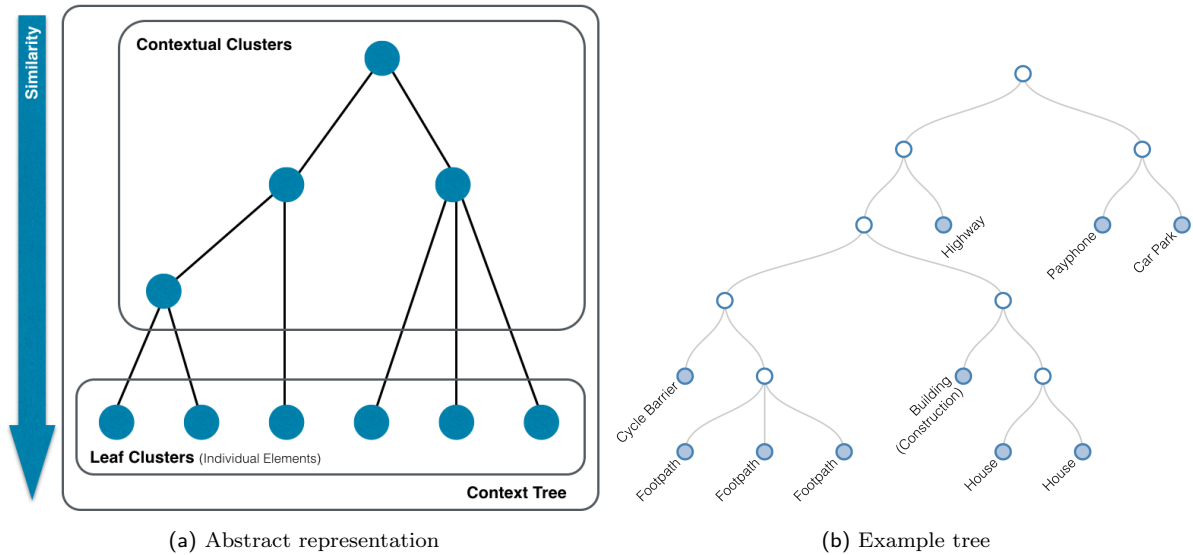

  \centering
  \begin{subfigure}[t]{0.49\linewidth}
  \includegraphics[width=1\linewidth]{figures/diagram_context_tree}
  \caption{Abstract representation}\label{fig:diagram_context_tree}
  \end{subfigure}
  \begin{subfigure}[t]{0.49\linewidth}   
  \includegraphics[width=1\linewidth]{figures/example_context_tree}
  \caption{Example tree}\label{fig:example_context_tree}
  \end{subfigure}
\caption{The context tree}\label{fig:context_tree}
\end{figure}

Towards the goal of identifying and representing contexts, Thomason \emph{et al.}~\citeyear{Thomason:ContextTrees} present the \emph{Context Tree}, a hierarchical data structure that summarises user contexts. Each leaf node of the tree represents a real-world feature or element that the user has likely interacted with, be it a specific building, area, or individual feature (e.g.\ a bench in a park). These individual elements are joined together through \emph{context nodes} that represent a context at a specific scale, where time spent within a context means that the user likely had similar aims or goals, and are identified by exploring time the user spends interacting with elements with similar properties, or elements that are interacted with in a similar manner. Figure~\ref{fig:diagram_context_tree} depicts the structure of a context tree, reproduced from~\cite{Thomason:ContextTrees}.

The procedure for generating a context tree augments geospatial trajectories with land usage data, filters these augmented trajectories to identify elements that the user was likely interacting with, and then clusters these interactions hierarchically using both features of the interactions and properties of the locations being interacted with.  This has advantages over existing work in that it assumes only the availability of geospatial trajectories at time of data collection and therefore does not rely on data from additional sensors.

Part of the work presented in this paper aims to convert the context tree into a predictive model which requires the use of hierarchical prediction techniques. Although hierarchical classification has been considered in other domains before, it still presents several challenges. The largest of these is a requirement for the model to understand the relationship between the nodes in the hierarchy. There are two primary ways of achieving this, either by maintaining the hierarchy and training a classifier at each node or level, the \emph{top-down} approach, or by constructing a new model that fully understands the hierarchical relationship~\cite{SillaJr:2011bm}. The latter category are classed as big-bang classifiers, and have been implemented using Bayesian models~\cite{Gopal:2012wv} and Markov networks~\cite{Rousu:2005hd}, while top-down classifiers typically employ support vector machines (SVM)~\cite{CesaBianchi:2006ew}.

\subsection{Geospatial Datasets}\label{sec:related:data}

Although geospatial trajectories are becoming increasingly common due to the now pervasive nature of location-aware smartphones, privacy concerns do present challenges for researchers. Available research datasets include Microsoft's GeoLife Trajectories~\cite{Zheng:2010uc} and Nokia's Mobile Data Challenge (MDC)~\cite{Kiukkonen:2010vm,Laurila:2012vk}. GeoLife was collected in such a way that participants were only rewarded for providing data when they were moving long distances, leading to large periods of missing data. MDC, on the other hand, aimed for continuous collection, but to protect privacy, periods around users' home locations have been obfuscated by truncating the longitude and latitude values, leading to artificial changes in variance. The MDC dataset contains data collected from the smartphones carried by nearly 200 users over a span of 2 years, and is the most appropriate of the publicly available datasets for evaluating the methods proposed in this paper. However, since it is not clear what impact these truncated periods would have, we opt to treat these periods as missing. 

Our primary evaluation data, however, comes from real-world data collected from 10 members of the University of Warwick, collected over 6 months, with a methodology aiming to replicate the MDC data collection process. This provides us with continuous data, without artificially missing periods, to conduct our evaluation. In addition, we use 10 users of the MDC data to verify the trends observed in the Warwick data. 


\section{Land Usage Extraction}\label{sec:augmentation}

The focus of this paper is using trajectories, collected by smartphones, and augmented with data from a land usage dataset to accurately identify the building or other real-world geographical feature that a person was interacting with at any given time. From this, the identification of meaningful locations becomes the task of grouping together spans of time in which the same element was interacted with. The benefits of this approach, instead of solely using geospatial trajectories as in existing approaches, is that it better captures the relationship between the identified locations and the real-world, with locations being representative in terms of shape, location and properties of the places actually interacted with by the user. 

This \emph{land usage extraction} procedure is designed to be able to replace existing location extraction approaches, where real-world elements replace the arbitrary clusters that would have previously been used. The procedure itself is based on the augmentation procedure used to generate context trees~\cite{Thomason:ContextTrees}. The primary difference, however, is that while the augmentation procedure for context trees allows an arbitrary number of land usage elements to be associated with each trajectory point, this process places an upper bound on both the size of elements and the number of elements that can be associated with a trajectory point. If, for example, this limit were set to 1, the resultant augmented trajectories would be synonymous with extracted locations, while higher values for multiple elements associated with each point. 

\subsection{Element Extraction} 

\begin{figure}[t]
  \centering
  \includegraphics[width=0.85\linewidth]{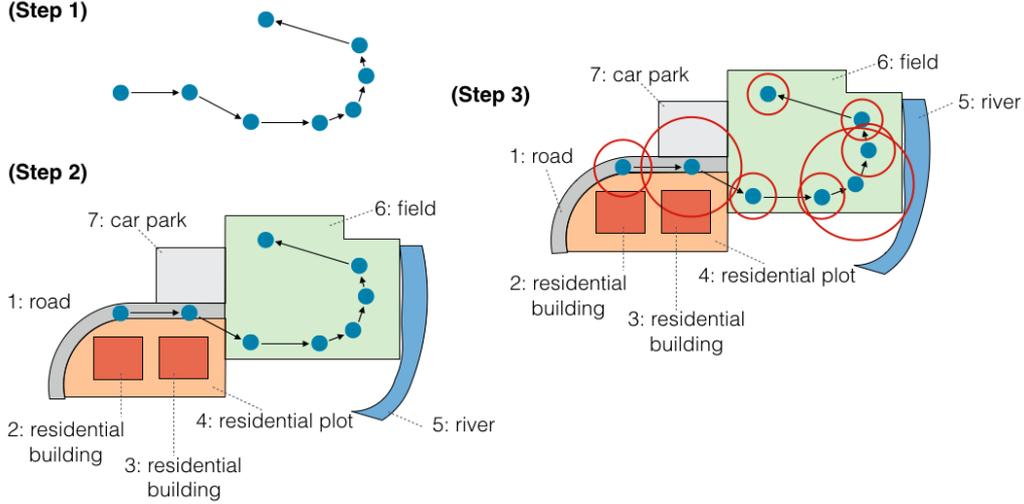}
  \caption{Extraction Procedure}\label{fig:diagram_extraction_1}
\end{figure}

A geospatial trajectory relates the location of an individual, entity or device to specific times. For this work, we assume that trajectories are temporally ordered and consist of longitude, latitude and timestamp as well as accuracy, measured in metres:
$$T = (\{t_1, lat_1, lng_1, acc_1\}, \{t_2, lat_2, lng_2, acc_2\},..., \{t_n, lat_n, lng_n, acc_n\})$$

This work requires both trajectories and land usage data, which are assumed to be sets of \emph{entities} with associated information. Each entity represents a real-world object, feature or area, such as an individual post box, building or farm. Elements are assumed to have a collection of geographical coordinates that represent their location and shape, along with a set of `key:value' pairs that describe its properties. For example, a house may have the tag `building:residential'.

The procedure for the extraction of relevant land usage elements, modified from~\cite{Thomason:ContextTrees}, is shown in Figure~\ref{fig:diagram_extraction_1}, where a raw geospatial trajectory (Step 1) enters the system and is overlaid on the land usage dataset (Step 2). The reported accuracy of each trajectory point is then used as a radius to consider (Step 3), such that all elements smaller than a specified size that are partially or wholly within the radius are stored alongside the original point (Table~\ref{tab:filtering:extracted}). This is achieved by processing each trajectory point in turn automatically until an augmented trajectory is formed. Elements associated with each point are then subjected to a filtering procedure, which identifies the $n$ most likely elements to have been being interacted with by the user. Once this has been completed, \emph{interactions} with elements are extracted by identifying when elements started and stopped being interacted with, consistent with the \emph{visits} in existing works. The example from Figure~\ref{fig:diagram_extraction_1} is finished with the summary shown in Table~\ref{tab:filtering:summarised}. This procedure differs from the original context tree generation process by introducing a parameter \textit{maxradius}, which specifies the maximum size of element to consider, preventing very large areas or groups of elements from being identified. Additionally, the parameter $n$ specifies the maximum number of land usage elements that can be associated with any trajectory point.

\begin{table}[t]
\centering
\parbox{.4\linewidth}{
\caption{Extracted data}\label{tab:filtering:extracted}
\begin{tabular}{l|ll}
\textbf{point} & \textbf{time} & \textbf{elements} \\ \hline
1              & 0             & \{1,2,4\}         \\
2              & 1             & \{1,3,4,6,7\}     \\
3              & 4             & \{6\}             \\
4              & 7             & \{6\}             \\
5              & 10            & \{5,6\}           \\
...            & ...           & ...               
\end{tabular}
}
\parbox{.4\linewidth}{
\caption{Summarised visits}\label{tab:filtering:summarised}
\begin{tabular}{lll}
\textbf{times} & \textbf{elem} & \textbf{tags}    \\ \hline
0-1            & 1             & road:residential \\
2-17           & 6             & landuse:field    \\
...            & ...           & ...               
\end{tabular}
}
\end{table}

\subsection{Filtering}\label{sec:filtering}

Identifying which land usage elements were most likely to have been interacted with is the task of a filtering procedure, which considers a buffer of points, consisting of a single \emph{point under consideration}, along with the $\delta$ seconds of data that fall either side of it. Each land usage element within this buffer is then scored according to the number of points the element is associated with, the accuracy of these points, and temporal distance from the point under consideration:
$$ Score(e) = \sum_{p\in P_e}\left(\frac{1}{a_p} \times \left(1-\frac{tdist(p, p_c)}{\delta}\right)\right) \times |P_e| $$

\noindent where $P_e$ is the set of all points that are associated with element $e$, $a_p$ is the accuracy value of point $p$, $p_c$ is the point under consideration, $\delta$ is the width of the buffer (i.e.\ the number of seconds from $p_c$ to consider) and $tdist(p_1,p_2)$ is the temporal distance between $p_1$ and $p_2$ (in seconds). This equation gives a higher score to elements associated with a large number of high accuracy points (where high accuracy is recorded as a small accuracy radius). From these scored elements, elements can be selected to be associated with each trajectory point, taking the $n$ elements with the highest scores. In rare cases, it is possible for two or more elements to share the same score, and in these instances, we select the elements closest to the trajectory point for consideration.

\subsection{Summarisation}

\begin{algorithm}[t] 
\caption{Summarisation Procedure}\label{alg:summarisation}
\algrenewcommand\alglinenumber[1]{\scriptsize #1:}
\algrenewcomment[1]{\(//\) #1}
\begin{algorithmic}[1]
\footnotesize

\State \textit{trajectory} $\gets (p_1, p_2,...)$ \Comment augmented trajectory
\State $t_{max} \gets$ 300 \Comment maximum time between consecutive points (seconds)
\State $d_{min} \gets$ 600 \Comment minimum visit duration (seconds)
\State \textit{elements} $\gets$ \textit{ElementStore} \Comment store of elements and their interactions
\State \textit{ongoing} $\gets \{\}$ \Comment stores start time of ongoing interactions until they are ended
\State \textit{previousTimestamp} $\gets$ \textit{p1}.\textit{timestamp}\\

\While{\textit{currentPoint} $\gets$ \textit{trajectory}.shift}\\
~~~~~~\Comment If too much time has past between points, end all ongoing interactions
  \If{(\textit{currentPoint}.\textit{timestamp} - \textit{previousTimestamp}) $> t_{max}$}
    \State \textit{toEnd} $\gets$ \textit{ongoing}
    \State \textit{toStart} $\gets$ \textit{currentPoint}.\textit{elements}
  \Else
    \State \textit{toEnd} $\gets$ (\textit{ongoing} - \textit{currentPoint}.\textit{elements}) \Comment End finished interactions
    \State \textit{toStart} $\gets$ (\textit{currentPoint}.\textit{elements} - \textit{ongoing}) \Comment Start interactions for new elements
  \EndIf\\
~\\

~~~~~\Comment Store interactions that are long enough
  \While{\textit{element} $\gets$ \textit{toEnd}.pop}
    \If{(\textit{previousTimestamp} - ongoing[element]) $> d_{min}$} 
      \State \textit{elements}.addInteraction(\textit{element}, \{start: ongoing[element], end: previousTimestamp\})
    \EndIf
    \State \textit{ongoing}.delete(\textit{element})
  \EndWhile\\
~\\

~~~~~\Comment Mark the start time of new interactions
  \While{\textit{element} $\gets$ \textit{toStart}.pop}
    \State \textit{ongoing}[\textit{element}] = \textit{currentPoint}.\textit{timestamp}
  \EndWhile\\

  \State \textit{previousTimestamp} $\gets$ \textit{currentPoint}.\textit{timestamp}

\EndWhile
\State \Return \textit{elements}

\end{algorithmic}
\end{algorithm}

Summarising the augmented trajectories into interactions is achieved through one-dimensional clustering that simply identifies neighbouring points that share the same land usage element, as shown in Algorithm~\ref{alg:summarisation}. This procedure requires two parameters: $t_{max}$, which prevents periods of missing data from being included in an interaction by specifying the maximum amount of time that can exist between neighbouring points before the interaction is split, and $d_{min}$, the minimum duration, required for an interaction to be stored. Upon completion, the procedure outputs a set of land usage elements that contain information about the element in the form of tags and coordinates, but also a set of times during which the user was interacting with the element. In cases where $n = 1$, i.e.\ the maximum number of land usage elements that can be associated with a trajectory point is 1 and, consequently, interactions cannot overlap, these interactions with elements are interchangeable with visits to locations as used in existing work.


\section{The Predictive Context Tree}\label{sec:pct}

Once the augmentation, filtering and summarising procedures have been completed, we are left with a record of the land usage elements interacted with by a particular user. These interactions become the basis for contextual clustering, forming a \emph{Context Tree} (described in Section~\ref{sec:related:contexttree}, and depicted earlier in Figure~\ref{fig:diagram_context_tree}). In this tree structure, leaf nodes represent individual land usage elements and nodes further up the hierarchy represent contextual clusters that contain all nodes below them. A small example context tree is shown in Figure~\ref{fig:example_context_tree}.

The process of converting a Context Tree into a Predictive Context Tree is the subject of the remainder of this section. As the Context Tree is a hierarchical structure, it contains a vast amount of information pertaining to the relationships between nodes, and so it is desirable to conserve this information. For this work, we opt to maintain this hierarchical structure and train the Context Tree as a hierarchical \emph{top-down} predictor (as discussed in Section~\ref{sec:related:contexttree}). The \emph{Predictive Context Tree} is an extension to the Context Tree data structure that is capable of both summarising a user's historical contexts as well as predicting their future context as a classification model. Initial Context Trees are trained according to the procedure outlined in~\cite{Thomason:ContextTrees} in combination with the augmentation procedure presented in Section~\ref{sec:augmentation}. The primary difference is that this method allows us to limit the size and number of land usage elements that can be associated with each trajectory point, for example limiting it to 1 element per point mirrors location extraction techniques commonly used in existing work (e.g.~\cite{Ashbrook:2003bi,Montoliu:2010bh}).

Once constructed, each node of the Context Tree, except the root node, becomes a binary classifier tasked with answering the question ``does the instance belong in the subtree rooted at this node?''. Overall classification of an instance occurs by starting at the root node and requesting a classification from each of the root's children, selecting children to follow based on a criteria until a final classification is reached, again determined by a criteria. The goals of prediction will determine the criteria:

\begin{figure}[t]
  \centering
  \includegraphics[width=0.7\linewidth]{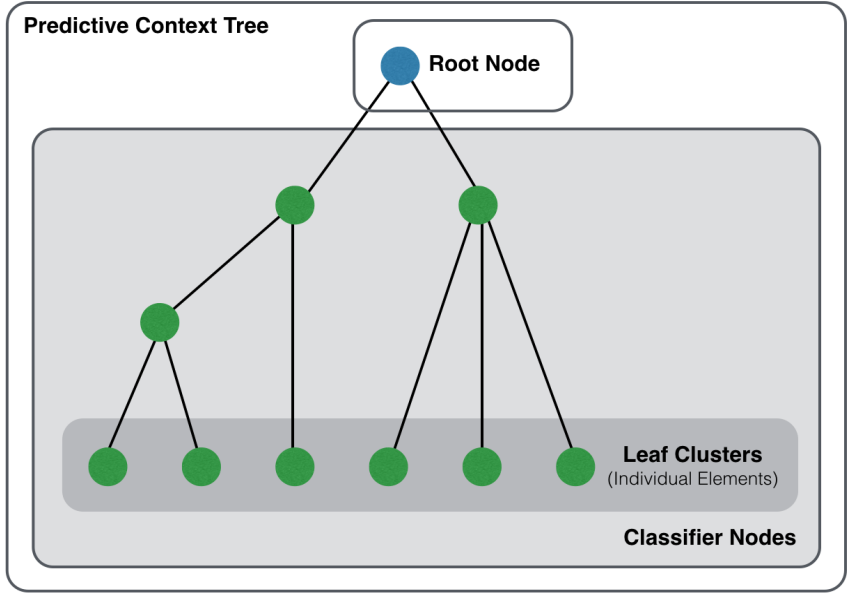}
  \caption{Abstract representation of a Predictive Context Tree}\label{fig:predictive_context_tree}
\end{figure}

\begin{figure}[t!]
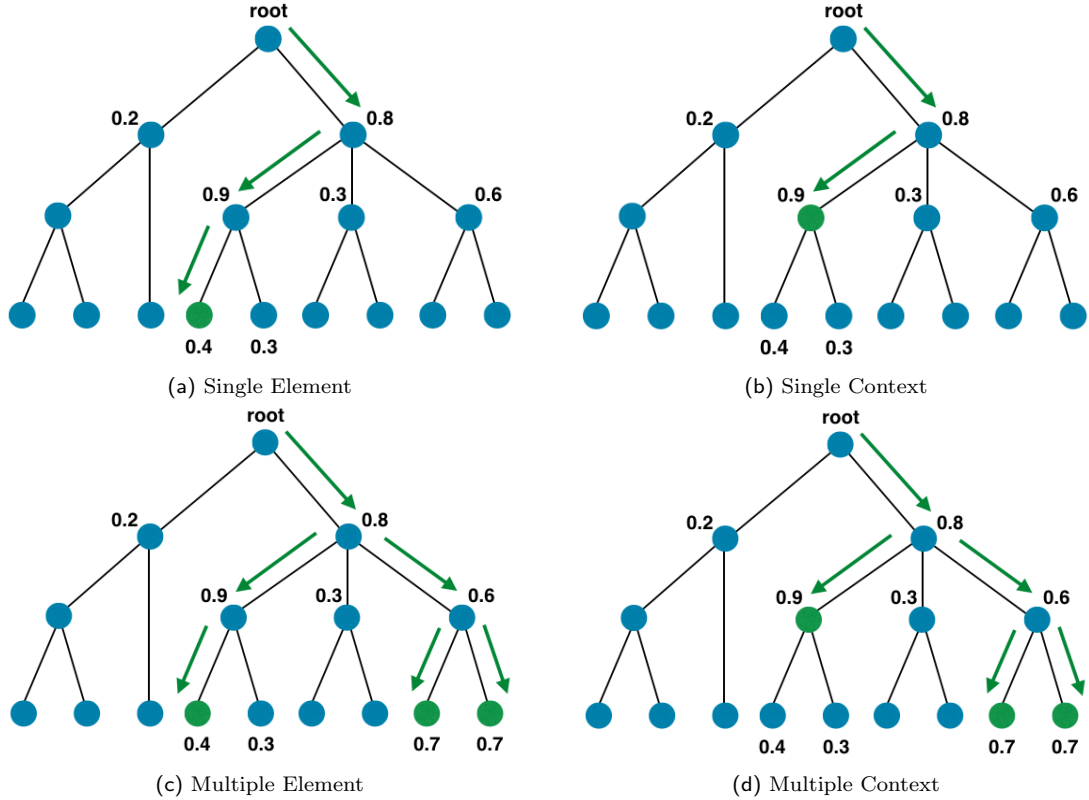

  \centering
  \footnotesize
  \begin{subfigure}[b]{0.47\linewidth}
    \centering
    \includegraphics[width=0.9\linewidth]{figures/pred-forced-leaf} 
    \caption{Single Element}\label{fig:pred-forced-leaf}
  \end{subfigure}
  \begin{subfigure}[b]{0.47\linewidth}
    \centering
    \includegraphics[width=0.9\linewidth]{figures/pred-context-single}
    \caption{Single Context}\label{fig:pred-context-single}
  \end{subfigure}
  \begin{subfigure}[b]{0.47\linewidth}
    \centering
    \includegraphics[width=0.9\linewidth]{figures/pred-multi-leaf} 
    \caption{Multiple Element}\label{fig:pred-multi-leaf}
  \end{subfigure}
  \begin{subfigure}[b]{0.47\linewidth}
    \centering
    \includegraphics[width=0.9\linewidth]{figures/pred-context-multi}
    \caption{Multiple Context}\label{fig:pred-context-multi}
  \end{subfigure}
  \caption{Classification methods for Predictive Context Trees. Classification begins at the root node and selects which children to follow based on the output of their binary classifiers, with different selection schemes}\label{fig:pred-approaches}
\end{figure}

\begin{itemize}
\item \textbf{Single element}: When predicting specific land usage elements, the predictor must return a leaf node, achieved by following the child with the highest confidence at each stage. This process is shown in Figure~\ref{fig:pred-forced-leaf}.
\item \textbf{Single context}: When predicting contexts instead of land usage elements, there is no requirement for a prediction to reach a leaf node. Figure~\ref{fig:pred-context-single} shows the procedure for single context prediction, where at each node, the child with the highest confidence is selected providing that the confidence is at least $T_s$. If no child has confidence of at least $T_s$, the current node is returned as the class label.
\item \textbf{Multiple element}: When trained on augmented trajectories that allow more than one land usage element to be associated with each trajectory point, PCTs can be used to predict the multiple land usage elements that will be interacted with next, unlike \emph{single element} which is limited to 1. This is achieved at each stage by following all classifiers that return confidence $> T_s$ if such exist, otherwise taking the one with the highest confidence, as shown in Figure~\ref{fig:pred-multi-leaf}.
\item \textbf{Multiple context}: Again, this technique aims to predict multiple elements, but when confidence is low in specific elements, contexts or combinations of elements and contexts can be predicted instead. Predictions are made by taking all children of a node whose confidence is above $T_s$, and returning the node when no child fulfils this criteria, as shown in Figure~\ref{fig:pred-context-multi}.
\end{itemize}

\emph{Single element} and \emph{multiple element} are examples of \emph{mandatory leaf node} prediction, in that the predictive model is required to return only leaf nodes from the tree. Similarly, \emph{multiple element} and \emph{multiple context} are examples of \emph{hierarchical multi-label} classifiers in that they can return more than one class label to a given test instance~\cite{SillaJr:2011bm}, although they can still return leaf nodes (i.e.\ elements) if confidence is high.

\subsection{Training a PCT}\label{sec:pct:training}

Training a PCT takes a set of \emph{instances} as the \emph{training set}, with known class labels. The PCT can be used for \emph{next-location} or \emph{next-context} prediction where the class label is an identifier pointing to the next location or context of the user after finishing with their \emph{current} context or location. In contrast, \emph{future-location} or \emph{future-context} prediction may generate training instances for each time step and the class label is simply which location or context the user was in during that time window. These instances are fed into each classifier in turn, with the class variable modified to become binary in the following ways:

\begin{itemize}
\item If the instance's class represents this node, it is used as a {\tt positive} training example
\item If the class represents a node in the subtree rooted at this node, it is {\tt positive}
\item If the class represents a sibling of this node, or a descendant of one, it is {\tt negative}
\item If the class represents an ancestor of this node, it is a {\tt negative} example
\item If the class represents any other node, it is ignored and not used for training
\end{itemize}

\begin{figure}
	\centering
	\includegraphics[width=0.42\linewidth]{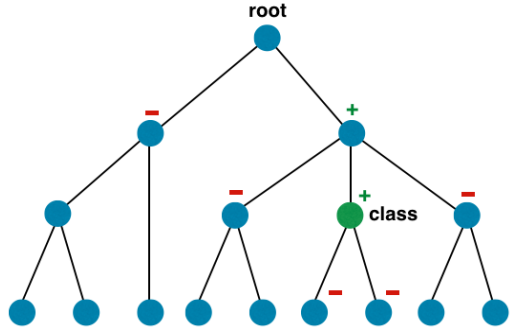}
	\caption{Example of how a training instance is treated by each classifier when the class label is associated with the node labelled `class'. All nodes labelled with `+' treat this instance as a positive example, nodes labelled `-' treat it as negative, while nodes without a label ignore this instance for training.}\label{fig:training-tree}
\end{figure}

Figure~\ref{fig:training-tree} shows how each node treats a particular instance. Training through this process ensures that the hierarchical links between elements and contexts are learnt by the PCT, as each node's classifier is trained to return {\tt yes} if the instance belongs to itself or one of its descendants, or {\tt no} if the instance belongs to a sibling or one of their descendants (i.e. following this particular child would be a mistake). Other nodes are ignored because they do not relate to the current problem.

Each node can now be trained as a binary classifier using standard techniques, such as decision trees or k-nearest neighbour approaches. However, using support vector machines, which are tailored to the task of binary classification, and have been shown to be applicable to the specific area of location prediction, are likely to be a good candidate here~\cite{Frohlich:2005fn,Wang:2012ta}.


\section{Experimental Setup}\label{sec:methodology}
The evaluation begins by considering the predictive accuracies that can be achieved when using existing location extraction techniques to identify locations visited by a user as the basis for prediction. We then consider the effects of replacing location extraction with element extraction (as presented in Section~\ref{sec:augmentation}). Finally, we use the PCT (presented in Section~\ref{sec:pct}) to replace the standard machine learning techniques as predictors. This section provides details of the experiments performed to evaluate the techniques proposed in this paper, with results presented in Section~\ref{sec:results}.

\subsection{Data}\label{sec:meth:data}
For this work, we use both geospatial trajectories collected from 10 members of the University of Warwick over a period of 6 months in 2014, and data from 10 users of the MDC dataset~\cite{Kiukkonen:2010vm,Laurila:2012vk} for evaluation. As we are unable to publish the Warwick data, for privacy and consent reasons, we use the MDC data for comparative purposes, despite the drawbacks discussed in Section~\ref{sec:related:data}. Land usage data is provided by OpenStreetMap (OSM)\footnote{\url{https://openstreetmap.org/}}, a community-maintained map of the world that contains information pertaining to real-world entities, including their geographical coordinates and a set of tags that describe the entity. 

\subsection{Location Prediction}\label{sec:meth:prediction}
The first stage of our evaluation considers location extraction and prediction using existing techniques, to provide a baseline for comparison. As discussed in Section~\ref{sec:related}, identifying visits, or \emph{interactions} as we will refer to them, is typically achieved using three thresholds: distance and time, which specify the maximum size and minimum duration, and $t_{max}$, the maximum time between two consecutive points to consider them part of the same interaction. Although less widely used than thresholding, the GVE algorithm has also been proposed for identifying such interactions, with the advantage that it can better handle noise in the data, and so we use both Thresholding and GVE for this task. We set the maximum distance parameter for Thresholding to 50m, to aim to extract locations no larger than a building. A value for $t_{max}$ of 1 hour allows for short periods of missing data, but will prevent longer periods from being included in interactions where the user may have left and returned some time later. For the MDC data, where longer periods of missing data are expected, we ignore this parameter. Finally, the time parameter, equivalent to the $d_{min}$ parameter in land usage extraction, is left open and its impact explored as part of the evaluation. The parameters for GVE allow for tuning the algorithm, but do not map neatly to the real-world properties of the extracted interactions (e.g.\ size and duration). To get around this, and produce comparable results, we first extract interactions using Thresholding and then select parameters for GVE that extract locations of approximately the same size, using the simulated annealing based methodology proposed in~\cite{Thomason:2015po}, where the distance metric is taken to be the difference in average location sizes between the data clustered with Thresholding and that clustered with GVE.

In both cases, interactions are clustered into locations using DBSCAN with $minpts = 0$, i.e.\ a single interaction can be considered as a location, and $eps = 15\mathrm{m}$, ensuring that interactions must be within proximity to be considered as part of the same location. Once locations have been extracted, prediction is considered using established techniques: \emph{support vector machines} and \emph{hidden Markov models}, both of which have been demonstrated to achieve high predictive accuracies for location prediction~\cite{Akoush:2007ct,Bilurkar:2002cf,Thomason:2015po,Wang:2012ta}.

\subsection{Land Usage Extraction}
Trajectories augmented with land usage information are to be used both as a foundation for prediction using existing techniques, and as a basis for contextual clustering and prediction using the PCT. In order to produce a representative comparison, parameters are selected that aim to mirror the extracted locations as best as possible. To this end, the maximum element size is constrained to be 50m across, $t_{max} = 1\mathrm{hr}$ is specified for the Warwick dataset (and ignored for the MDC data), $n$, the maximum number of elements to be associated with a trajectory point, is set to 1, and the same values of $d_{min}$ as used to extract locations (Section~\ref{sec:meth:prediction}) are used for exploring its impact on predictive accuracy. The only additional parameter required by this procedure is $\delta$, specifying the width of the buffer to consider during the filtering stage of trajectory augmentation~\cite{Thomason:2016gv}. We set $\delta = 5\mathrm{min}$ for this task, a value selected empirically that produces representative results. Predicting over land usage elements is performed using the same machine learning techniques as in extracted locations, namely support vector machines and hidden Markov models. 

\subsection{Instance Generation}\label{sec:instance_generation}
For both the extracted location and land usage datasets, training instances must be generated. This is achieved by selecting interactions with locations or extracted features that last longer than $d_{min}$ minutes. Instances are then generated by summarising interactions into a set of features: \emph{day of year}, \emph{day of week}, \emph{start hour}, \emph{start minute}, \emph{duration}, \emph{current identifier (element or location)}, and \emph{class} (next identifier).

\subsection{Multi-element Land Usage}
In addition to the land usage extraction procedure already discussed, we explore the potential of allowing overlapping and parallel interactions where multiple land usage elements may be associated with each trajectory point. These may be useful if, for example, a person is interacting with a building that is contained within a larger building (e.g.\ a shop in a shopping centre), and so we also utilise the land usage extraction procedure with different values of $n$. For \emph{multi-element land usage extraction}, we set the maximum element size to be 100m, instead of 50m for \emph{single-element}, allowing larger buildings to also be extracted. Generating training instances for these datasets uses the same features as before, however the class label becomes the set of elements that the user interacts with next. This is defined as taking the next interaction in the dataset, selecting all other interactions that overlap, and combining the identifiers of all such elements into a single string value that represents the complete set of elements.

\subsection{Constructing Predictive Context Trees}\label{sec:data:augmenting}
Context trees are constructed from the land usage datasets, both single and multi, according to the procedure presented in~\cite{Thomason:ContextTrees}, and using the Hybrid Contextual Distance (HCD) metric with $\lambda = 0.5$, an empirically selected value that produces representative results. The HCD metric is a similarity measure that balances semantic and feature similarities into a single score used for clustering context trees, where $\lambda$ specifies the weighting towards semantic similarity.

The task now becomes that of converting the generated Context Trees into Predictive Context Trees and evaluating the predictive ability of such a hierarchical model. Each non-root node in the context tree is trained as a binary classifier using a support vector machine (SVM) with the modification of instances as described in Section~\ref{sec:pct:training}. 

\subsection{Evaluating Predictions}\label{sect:meth:evaluation}
For all location and element prediction approaches (extracted location, land usage, and single element PCT), the training data's class label represents the next extracted location or land usage element that the user interacted with. Evaluating the correctness of a prediction can simply be performed by comparing the output of the predictor against the known class, referred to as an \emph{element correct} prediction. For \emph{context} prediction, in some cases the PCT will return a leaf node which can then be compared to see if it is \emph{element correct}. In other cases, a non-leaf node will be returned which requires the introduction of the notion of \emph{context correctness}:

\begin{definition}\label{def:contextually_correct}
A prediction is \textbf{context correct} if the node represented by the predicted class label is an ancestor of the actual class node.
\end{definition}

For trees constructed over multi-element land usage datasets, predictions take the form of a set of elements or contexts, and so we require evaluative methods to differentiate between these predictions. For this, we define several tests, applied in order:

\begin{description}
\item[\textbf{Fully element correct}] The set of predicted land usage elements matches the set of actual elements exactly
\item[\textbf{Fully context correct}] Every member in the set of actual elements is represented in the predicted set either by itself or an ancestor in the tree. Additionally, every element in the predicted set is either contained within, or an ancestor of, at least one element in the actual set
\item[\textbf{Partial element correct}] Some elements were correctly predicted: the union of the predicted and actual sets is non empty
\item[\textbf{Partial context correct}] Some contexts were correctly predicted: the union of the predicted set and the set of all ancestors of members of the actual set is non empty
\item[\textbf{Incorrect}] There is no overlap between the predicted and actual set, or the predicted and ancestors of members of the actual set
\end{description}


\section{Results}\label{sec:results}

\begin{figure}[t]
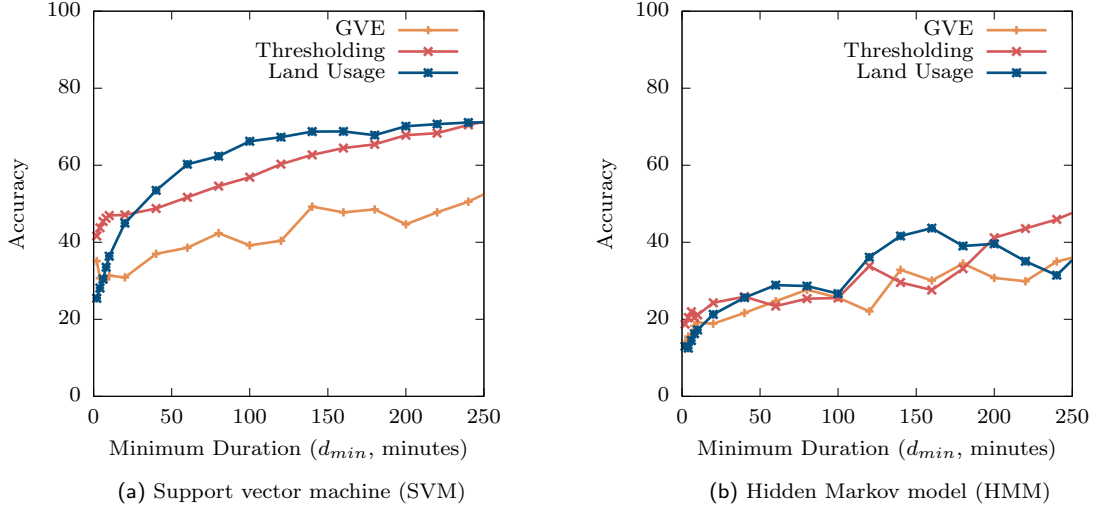

  \centering
  \footnotesize
  \begin{subfigure}[t]{0.48\textwidth}
  \begin{gnuplot}[terminal=epslatex, terminaloptions={size 2.7,2.5 font 8}]
    set key right
    #set key spacing 1.5
    #set key width 3

    set ylabel "Accuracy"
    set xlabel "Minimum Duration ($d_{min}$, minutes)"
    set yrange [0:100]
    set xrange [0:250]
 
    #set arrow from 30,graph(0,0) to 30,graph(1,1) nohead dashtype 2 lc rgb 'red' lw 4

    plot 'data/dmin-trajectory.dat' using ($1/60):4 t "GVE" with linespoints lw 4 lc "#e58e53",\
    '' using ($1/60):5 t "Thresholding" with linespoints lw 4 lc "#d75959",\
    'data/dmin-landusage.dat' using ($1/60):3 t "Land Usage" with linespoints lw 4 lc "#065183" #$
  \end{gnuplot}
  \caption{Support vector machine (SVM)}\label{fig:res:dmin:svm}
\end{subfigure}
\begin{subfigure}[t]{0.48\textwidth}
  \begin{gnuplot}[terminal=epslatex, terminaloptions={size 2.7,2.5 font 8}]
    set key right#above opaque invert
    #set key spacing 1.5
    #set key width 3

    set ylabel "Accuracy"
    set xlabel "Minimum Duration ($d_{min}$, minutes)"
    set yrange [0:100]
    set xrange [0:250]
 
    #set arrow from 30,graph(0,0) to 30,graph(1,1) nohead dashtype 2 lc rgb 'red' lw 4

    plot 'data/dmin-trajectory.dat' using ($1/60):2 t "GVE" with linespoints lw 4 lc "#e58e53",\
    '' using ($1/60):3 t "Thresholding" with linespoints lw 4 lc "#d75959",\
    'data/dmin-landusage.dat' using ($1/60):2 t "Land Usage" with linespoints lw 4 lc "#065183"#$
  \end{gnuplot}
  \caption{Hidden Markov model (HMM)}\label{fig:res:dmin:hmm}
\end{subfigure}
\caption{Predictive accuracy for different minimum interaction durations using existing location extraction and prediction techniques as well as the proposed land usage identification scheme}\label{fig:res:dmin}
\end{figure}

\begin{figure}[ht!]
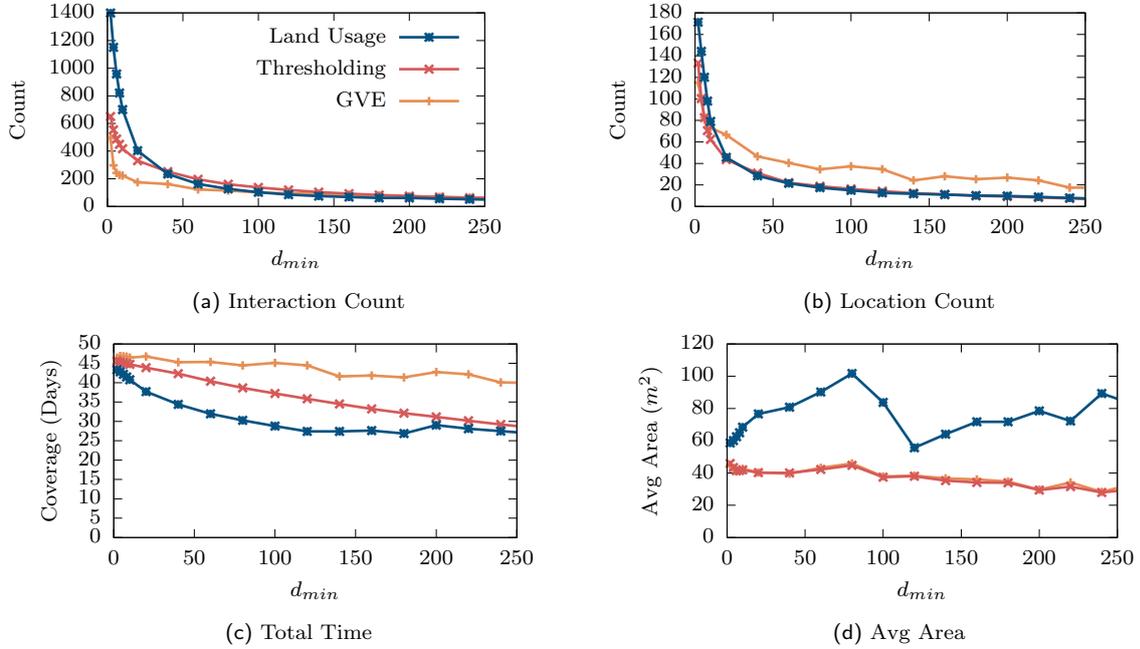

  \centering
  \footnotesize

  \begin{subfigure}[b]{0.49\linewidth}
  \begin{gnuplot}[terminal=epslatex, terminaloptions={size 2.7,1.5 font 8}]
    set key right opaque invert
    set key spacing 1.5
    #set key width 3

    set ylabel "Count"
    set xlabel "$d_{min}$"
    #set yrange [0:100]
    set xrange [0:250]
 
    plot 'data/location-summaries-visit-count.dat' using ($1/60):2 t "GVE" with linespoints lw 4 lc "#e58e53",\
    '' using ($1/60):3 t "Thresholding" with linespoints lw 4 lc "#d75959",\
    '' using ($1/60):4 t "Land Usage" with linespoints lw 4 lc "#065183"#$

  \end{gnuplot}
  \caption{Interaction Count}\label{fig:res:location_summary:visit_count}
\end{subfigure}
\begin{subfigure}[b]{0.49\linewidth}
  \begin{gnuplot}[terminal=epslatex, terminaloptions={size 2.7,1.5 font 8}]
    set key off #above opaque invert
    #set key spacing 1.5
    #set key width 3

    set ylabel "Count"
    set xlabel "$d_{min}$"
    #set yrange [0:100]
    set xrange [0:250]
 
    plot 'data/location-summaries-location-count.dat' using ($1/60):2 t "GVE" with linespoints lw 4 lc "#e58e53",\
    '' using ($1/60):3 t "Thresholding" with linespoints lw 4 lc "#d75959",\
    '' using ($1/60):4 t "Land Usage" with linespoints lw 4 lc "#065183"#$

  \end{gnuplot}
  \caption{Location Count}\label{fig:res:location_summary:location_count}
\end{subfigure}
\begin{subfigure}[b]{0.49\linewidth}
  \centering
  \footnotesize
  \begin{gnuplot}[terminal=epslatex, terminaloptions={size 2.7,1.5 font 8}]
    set key off #above opaque invert
    #set key spacing 1.5
    #set key width 3

    set ylabel "Coverage (Days)"
    set xlabel "$d_{min}$"
    set yrange [0:]
    set xrange [0:250]
 
    plot 'data/location-summaries-total-time.dat' using ($1/60):($2/(60*60*24)) t "GVE" with linespoints lw 4 lc "#e58e53",\
    '' using ($1/60):($3/(60*60*24)) t "Thresholding" with linespoints lw 4 lc "#d75959",\
    '' using ($1/60):($4/(60*60*24)) t "Land Usage" with linespoints lw 4 lc "#065183"

  \end{gnuplot}
  \caption{Total Time}\label{fig:res:location_summary:total_time}
\end{subfigure}
\begin{subfigure}[b]{0.49\linewidth}
  \centering
  \footnotesize
  \begin{gnuplot}[terminal=epslatex, terminaloptions={size 2.7,1.5 font 8}]
    set key off #above opaque invert
    #set key spacing 1.5
    #set key width 3

    set ylabel "Avg Area ($m^2$)"
    set xlabel "$d_{min}$"
    set yrange [0:]
    set xrange [0:250]
 
    plot 'data/avg-location-area-traj.dat' using ($1/60):2 t "GVE" with linespoints lw 4 lc "#e58e53",\
    '' using ($1/60):3 t "Thresholding" with linespoints lw 4 lc "#d75959",\
    'data/avg-location-area-lu.dat' using ($1/60):4 t "Land Usage" with linespoints lw 4 lc "#065183"#$

  \end{gnuplot}
  \caption{Avg Area}\label{fig:res:location_summary:avg_area}
\end{subfigure}
\caption{Summaries of properties of extracted locations and identified elements}\label{fig:res:location_summary}
\end{figure}

Evaluating the performance of the PCT requires a baseline from existing approaches. Figure~\ref{fig:res:dmin} provides insight into the accuracies that can be expected from existing prediction models (support vector machines and hidden Markov models) when predicting over locations extracted using existing clustering techniques and land usage elements extracted using the procedure presented in this paper. The figure demonstrates that increasing $d_{min}$ leads to predictions of higher accuracy, as $d_{min}$ controls the minimum duration of an interaction to consider, with shorter interactions being ignored as noise. A larger value for $d_{min}$ only considers locations at which the user has spent significant amounts of time, thereby making predictions more accurate with fewer possible locations the user will visit. Additionally, the results show that support vector machines outperform hidden Markov models in all cases. Of most relevance, however, is the relative performance of the predictors operating over land usage elements when compared to those operating over extracted locations. For short interaction durations, extracted locations provide the foundation that affords more accurate predictions, but as $d_{min}$ is increased beyond 30 minutes, extracted land usage elements provide the better foundation, as demonstrated by the higher predictive accuracies observed. 

Figure~\ref{fig:res:location_summary} shows summaries of different properties of the extracted location and identified element datasets for the different values of $d_{min}$, to understand how comparable the results are. Specifically, Figure~\ref{fig:res:location_summary:visit_count} shows the number of interactions present in each dataset (averaged across the 10 users), and Figure~\ref{fig:res:location_summary:location_count} shows the number of different extracted locations or identified land usage elements associated with these interactions. Figure~\ref{fig:res:location_summary:total_time} shows the total time contained within these interactions, and Figure~\ref{fig:res:location_summary:avg_area} shows the average area of locations and land usage elements. The graphs show that higher values of $d_{min}$ lead to fewer interactions, locations or elements and a lower coverage of total time. The average area is least impacted by $d_{min}$, but also the most varied among the techniques. GVE and Thresholding have similar values, as parameters for GVE were selected specifically to extract locations with similar sizes to those identified through Thresholding, while the land usage elements are consistently larger. Larger land usage elements are to be expected as extracted locations only consider regions where users actually spent time, while land usage elements consider entire buildings or features where a user may have only interacted with part of. Figure~\ref{fig:res:location_summary:location_count} also helps to understand why the predictive accuracy for GVE is lower than Thresholding (as shown in Figure~\ref{fig:res:dmin}): for each value of $d_{min} > 30$, GVE consistently extracts more locations than Thresholding, which would result in more possible class labels and therefore more complex models, and consequently lower predictive accuracy.



\begin{figure}[t]
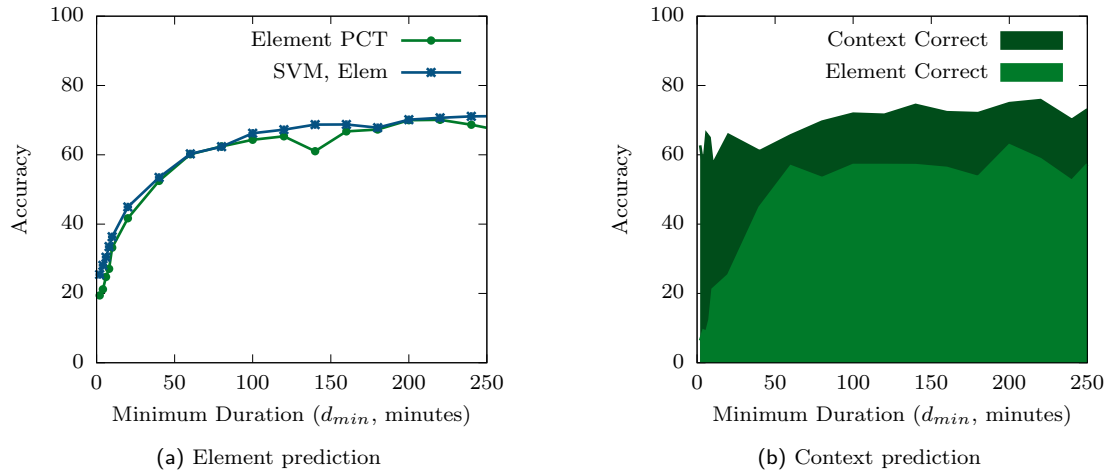

  \centering
  \footnotesize

  \begin{subfigure}[b]{0.49\linewidth}
  \centering
  \footnotesize
  \begin{gnuplot}[terminal=epslatex, terminaloptions={size 2.7,2.3 font 8}]
    set key top opaque
    set key spacing 1.5
    set key width 3

    set ylabel "Accuracy"
    set xlabel "Minimum Duration ($d_{min}$, minutes)"
    set yrange [0:100]
    set xrange [0:250]
 
    #set arrow from 30,graph(0,0) to 30,graph(1,1) nohead dashtype 2 lc rgb 'red' lw 4

    plot 'data/tree-singleelement-dmin.dat' using ($1/60):($2+$3) t "Element PCT" with linespoints pt 7 lw 4 lc "#007a28",\
    'data/dmin-landusage.dat' using ($1/60):3 t "SVM, Elem" with linespoints pt 3 lw 4 lc "#065183"
    
  \end{gnuplot}
  \caption{Element prediction}\label{fig:res:singleelement}
  \end{subfigure}
  \begin{subfigure}[b]{0.49\linewidth}
  \centering
  \footnotesize
  \begin{gnuplot}[terminal=epslatex, terminaloptions={size 2.7,2.3 font 8}]
    set key top opaque
    set key spacing 1.5
    set key width 3

    set ylabel "Accuracy"
    set xlabel "Minimum Duration ($d_{min}$, minutes)"
    set yrange [0:100]
    set xrange [0:250]
 
    #set arrow from 30,graph(0,0) to 30,graph(1,1) nohead dashtype 2 lc rgb 'red' lw 4

    plot 'data/tree-singlecontext-dmin.dat' using ($1/60):($2+$3+$4) t "Context Correct" lw 4 lc "#004c19" with filledcurves x1,\
    '' using ($1/60):($2+$3) t "Element Correct" lw 4 lc "#007a28" with filledcurves x1,#$
  \end{gnuplot}
  \caption{Context prediction}\label{fig:res:singlecontext}
\end{subfigure}
\caption{Predictive accuracy when using the Predictive Context Tree for both element and context prediction}\label{fig:res:pct}
\end{figure}

\begin{figure}[t]
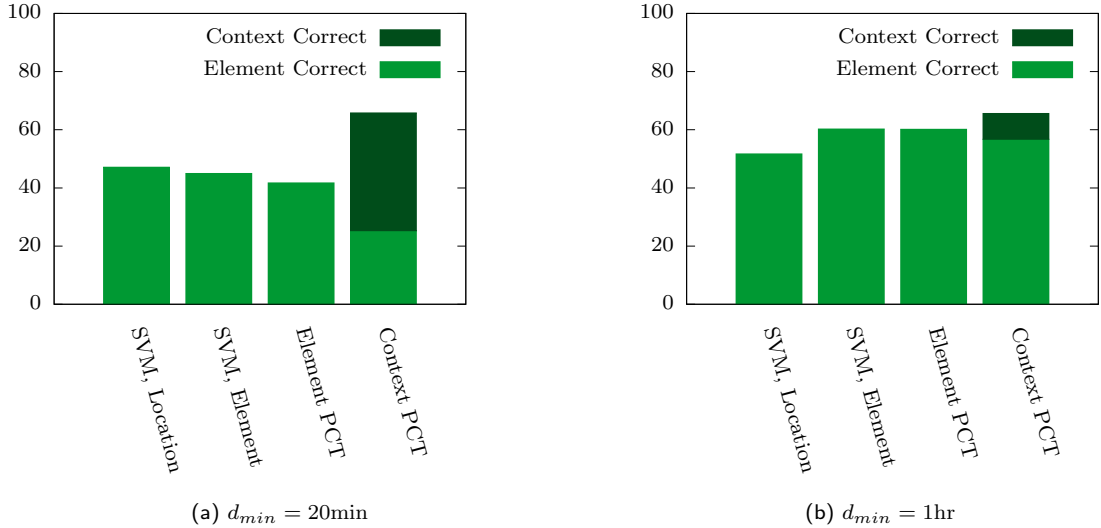

  \centering
  \footnotesize

  \begin{subfigure}[b]{0.49\linewidth}
    \begin{gnuplot}[terminal=epslatex, terminaloptions={size 2.7,2.6 font 8}]
    set key right invert #box opaque invert
    set key spacing 1.5
    # set key width -1

    set style data histograms
    set style histogram rowstacked

    set boxwidth 0.8
    set style fill solid

    set ylabel "" #"Percentage"
    set xlabel "" # Prediction Scheme"
    set yrange [0:100]

    set datafile missing "?"
    
    set xtic rotate by -75 scale 0 offset 0,-0.5

    plot 'data/1200-comparison.dat' using 2:xtic(1) lc '#009933' t 'Element Correct',\
    '' using 3 lc '#004c19' t 'Context Correct'

  \end{gnuplot}
  \caption{$d_{min} = 20\mathrm{min}$}\label{fig:res:comparison:1200}
\end{subfigure}
\begin{subfigure}[b]{0.49\linewidth}  \centering
  \footnotesize
  \vspace{5pt}
    \begin{gnuplot}[terminal=epslatex, terminaloptions={size 2.7,2.6 font 8}]
    set key right invert #box opaque invert
    set key spacing 1.5
    # set key width -1

    set style data histograms
    set style histogram rowstacked

    set boxwidth 0.8
    set style fill solid

    set datafile missing "?"

    set ylabel "" #"Percentage"
    set xlabel "" # Prediction Scheme"
    set yrange [0:100]
    
    set xtic rotate by -75 scale 0 offset 0,-0.5

    plot 'data/3600-comparison.dat' using 2:xtic(1) lc '#009933' t 'Element Correct',\
    '' using 3 lc '#004c19' t 'Context Correct'

  \end{gnuplot}
  \caption{$d_{min} = 1\mathrm{hr}$}\label{fig:res:comparison:3600}
\end{subfigure}
\caption{Comparison of predictive methods, where locations are extraction using the Thresholding technique}\label{fig:res:comparison}
\end{figure}

With comparative baselines in place, the task becomes that of understanding the relative performance of the PCT. Figure~\ref{fig:res:singleelement} shows the predictive accuracies achieved when using the PCT to predict which land usage element a user will interact with, the same task performed by the SVM in Figure~\ref{fig:res:dmin:svm}, and so the performance of the SVM predictor is shown as a comparison. The figure demonstrates that the PCT produces comparable predictive accuracies to SVM when predicting the next element a person will interact with. The PCT is also designed to predict contexts as well as elements, and the accuracies achieved for context prediction are shown in Figure~\ref{fig:res:singlecontext}. A comparison of the performance of all techniques are shown in Figure~\ref{fig:res:comparison} for $d_{min} = 20\mathrm{min}$ and $1\mathrm{hr}$, where location extraction was performed by \emph{Thresholding}, since it gives the highest predictive accuracies. For $d_{min} = 20\mathrm{min}$, predicting over extracted locations using SVMs provides the highest element correct accuracy (i.e.\ it is best able to predict the exact location or element to be interacted with), however, when allowing for contextual prediction, the PCT outperforms this existing approach. With larger values of $d_{min}$, such as $1\mathrm{hr}$, shown in Figure~\ref{fig:res:comparison:3600}, predicting over identified land usage elements, as proposed in this work, provides higher accuracies than predicting over extracted locations for both the existing SVM based approach, and the PCT. Again, allowing for contextual prediction, the PCT outperforms the other approaches.

\begin{figure}[t]
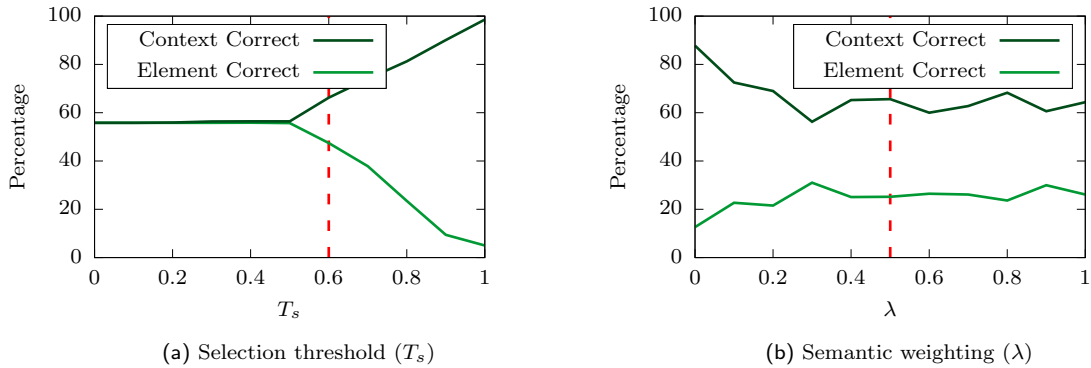

  \centering
  \footnotesize
  \begin{subfigure}[b]{0.49\linewidth}
  \begin{gnuplot}[terminal=epslatex, terminaloptions={size 2.7,1.75 font 8}]
    set key left box opaque invert
    set key spacing 1.5
    # set key width -1

    set ylabel "Percentage"
    set xlabel "$T_s$"
    set yrange [0:100]
    #set xrange [-0.05:3.65]
    
    #set xtic offset 2,0 #rotate by -45 scale 0

    set arrow from 0.6,graph(0,0) to 0.6,graph(1,1) nohead dashtype 2 lc rgb 'red' lw 4
    #set arrow from 0.005,graph(0,0) to 0.005,graph(1,1) nohead dashtype 0 lc rgb 'blue' lw 4

    plot 'data/followthreshold.dat' using 1:2 t 'Element Correct' with lines lw 4 lc '#009933' smooth unique,\
    '' using 1:($2+$3) t 'Context Correct' with lines lw 4 lc '#004c19' smooth unique
  \end{gnuplot}
  \caption{Selection threshold ($T_s$)}\label{fig:res:threshold}
  \end{subfigure}
  \begin{subfigure}[b]{0.49\linewidth}
  \begin{gnuplot}[terminal=epslatex, terminaloptions={size 2.7,1.75 font 8}]
    set key right box opaque invert
    set key spacing 1.5
    # set key width -1

    set ylabel "Percentage"
    set xlabel "$\\lambda$"
    set yrange [0:100]
    #set xrange [-0.05:3.65]
    
    #set xtic offset 2,0 #rotate by -45 scale 0

    set arrow from 0.5,graph(0,0) to 0.5,graph(1,1) nohead dashtype 2 lc rgb 'red' lw 4
    #set arrow from 0.005,graph(0,0) to 0.005,graph(1,1) nohead dashtype 0 lc rgb 'blue' lw 4

    plot 'data/lambda.dat' using 1:2 t 'Element Correct' with lines lw 4 lc '#009933' smooth unique,\
    '' using 1:($2+$3) t 'Context Correct' with lines lw 4 lc '#004c19' smooth unique
  \end{gnuplot}
  \caption{Semantic weighting ($\lambda$)}\label{fig:res:lambda}
\end{subfigure}
  \caption{Exploring the impact of the parameters $T_s$ and $\lambda$ on context prediction, where $d_{min} = 20\mathrm{min}$. The dashed red lines show the values used for previous figures.}\label{fig:res:context:parameters}
\end{figure}

The impact of two of the remaining parameters, $T_s$ and $\lambda$, for context prediction using the PCT are shown in Figure~\ref{fig:res:context:parameters}. $T_s$ is the \emph{Selection Threshold}, specifying the predictive confidence required to follow a node through to its child when traversing the context tree. A higher value for $T_s$ will mean that children are less likely to be followed, instead returning contextual predictions, leading to fewer element correct predictions, but more context correct ones. The \emph{Semantic Weighting} parameter, $\lambda$, is used when clustering context trees. A value of 0 uses only the feature similarity for determining contextual clusters, while a value of 1 uses only the semantic similarity~\cite{Thomason:ContextTrees}. When considering primarily feature similarity, the element prediction decreases in accuracy, but the context prediction increases. This indicates that the contexts identified are far less meaningful as they do not provide a good basis for predicting specific elements. Increasing $\lambda$ both increases the accuracy of element prediction and decreases the accuracy of context prediction. As more accurate element predictions are seen, the contexts become more representative and useful as an indicator of what the person is doing, leading to a more useful predictive model.

\begin{figure}[t]
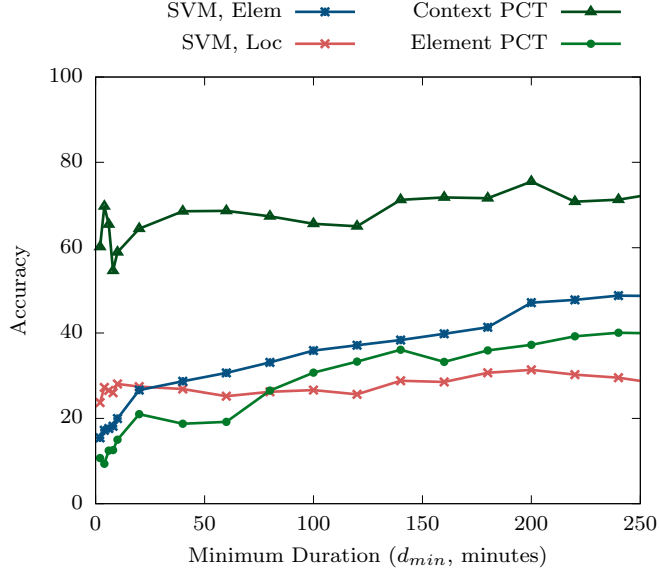

  \centering
  \footnotesize
  \begin{gnuplot}[terminal=epslatex, terminaloptions={size 3.5,3.05 font 8}]
    set key above opaque invert
    set key spacing 1.5
    set key width 3

    set ylabel "Accuracy"
    set xlabel "Minimum Duration ($d_{min}$, minutes)"
    set yrange [0:100]
    set xrange [0:250]
 
    #set arrow from 30,graph(0,0) to 30,graph(1,1) nohead dashtype 2 lc rgb 'red' lw 4

    plot 'data/dmin-trajectory-mdc.dat' using ($1/60):3 t "SVM, Loc" with linespoints lw 4 lc "#d75959" pt 2,\
    'data/dmin-landusage-mdc.dat' using ($1/60):3 t "SVM, Elem" with linespoints lw 4 lc "#065183" pt 3,\
    'data/tree-singleelement-dmin-mdc.dat' using ($1/60):($2+$3) t "Element PCT" with linespoints pt 7 lw 4 lc "#007a28",\
    'data/tree-singlecontext-dmin-mdc.dat' using ($1/60):($2+$3+$4) t "Context PCT" with linespoints pt 8 lw 4 lc "#004c19"#$
  \end{gnuplot}
  \caption{Predictive accuracies for the different techniques for the MDC dataset}\label{fig:res:mdc}
\end{figure}

As Section~\ref{sec:meth:data} discusses, the results presented so far have all come from the Warwick dataset as this has a high level of coverage. In order to demonstrate the applicability of the PCT to other data, we also use a subset of the MDC dataset for comparison, shown in Figure~\ref{fig:res:mdc}. Although the MDC data contains truncated latitude and longitude values in parts, the trends are consistent with previous results. Contextual PCT prediction outperforms the other approaches, with SVM and the PCT in element prediction mode performing similarly. Finally, when using SVMs over extracted locations, the accuracy starts slightly higher than over identified land usage elements (as with Figure~\ref{fig:res:dmin}), but with $d_{min} > 30$, land usage elements provide higher predictive accuracies.

\subsection{Pruning}

\begin{figure}[t]
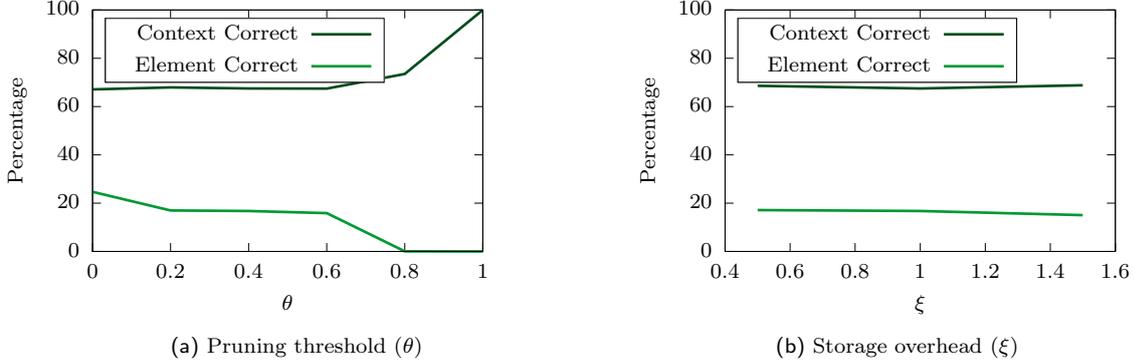

  \centering
  \footnotesize
  \begin{subfigure}[b]{0.49\linewidth}
  \begin{gnuplot}[terminal=epslatex, terminaloptions={size 2.7,1.75 font 8}]
    set key left box opaque invert
    set key spacing 1.5
    # set key width -1

    set ylabel "Percentage"
    set xlabel "$\\theta$"
    set yrange [0:100]
    #set xrange [-0.05:3.65]
    
    plot 'data/pruned-theta.dat' using 1:2 t 'Element Correct' with lines lw 4 lc '#009933' smooth unique,\
    '' using 1:($2+$3) t 'Context Correct' with lines lw 4 lc '#004c19' smooth unique
  \end{gnuplot}
  \caption{Pruning threshold ($\theta$)}\label{fig:res:pruned_theta}
  \end{subfigure}
  \begin{subfigure}[b]{0.49\linewidth}
  \centering
  \footnotesize
  \begin{gnuplot}[terminal=epslatex, terminaloptions={size 2.7,1.75 font 8}]
    set key left box opaque invert
    set key spacing 1.5
    # set key width -1

    set ylabel "Percentage"
    set xlabel "$\\xi$"
    set yrange [0:100]
    #set xrange [-0.05:3.65]

    plot 'data/pruned-xi.dat' using 1:2 t 'Element Correct' with lines lw 4 lc '#009933' smooth unique,\
    '' using 1:($2+$3) t 'Context Correct' with lines lw 4 lc '#004c19' smooth unique
  \end{gnuplot}
  \caption{Storage overhead ($\xi$)}\label{fig:res:pruned_xi}
\end{subfigure}
  \caption{Predictive accuracies from pruned context trees}\label{fig:res:pruned}
\end{figure}

The context tree generation procedure (\cite{Thomason:ContextTrees}), includes an approach to pruning superfluous nodes to reduce storage and processing requirements. This process takes two parameters: $\theta$ and $\xi$, where $\theta$ specifies a threshold between 0 and 1 for a node to be pruned, and $\xi$ assigns a storage overhead to each node in the tree. Figure~\ref{fig:res:pruned} shows the impact of these parameters on predictive accuracy. A larger value of $\theta$ (Figure~\ref{fig:res:pruned_theta}) leads to more nodes being removed from the tree, resulting in fewer element correct predictions (as the leaf nodes corresponding to the elements are removed), but an increased number of context correct predictions. The effect of $\xi$ is similar as its value is increased. A higher overhead assigned to storing a node makes it more likely that the node will be pruned. If using the context tree for predictive applications, through the PCT, these results indicate that a reduced size tree comes at a trade-off of reduced element correct predictive accuracy. 

\subsection{Multi-element Prediction}

\begin{figure}[t]
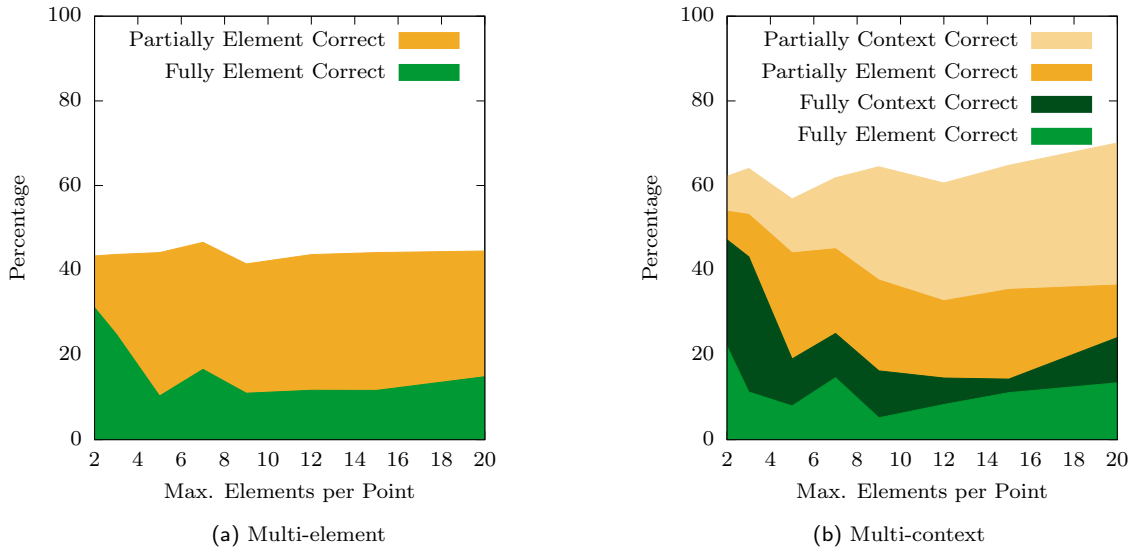

  \centering
  \footnotesize
  \begin{subfigure}[b]{0.49\linewidth}
  \begin{gnuplot}[terminal=epslatex, terminaloptions={size 2.7,2.7 font 8}]
      set key right #above box opaque
      set key spacing 1.5
      # set key width -1

      #set style data histograms
      #set style histogram rowstacked

      #set boxwidth 0.5
      #set style fill solid

      set ylabel "Percentage"
      set xlabel "Max. Elements per Point"
      set yrange [0:100]
      #set xrange [-0.05:3.65]
      
      #set xtic offset 2,0 #rotate by -45 scale 0

      plot 'data/tree-multielement-dmin.dat' using 1:($2+$3) t 'Partially Element Correct' with filledcurves x1 lc '#F1AA24',\
      '' using 1:($2) t 'Fully Element Correct' with filledcurves x1 lc '#009933' #$
  \end{gnuplot}
  \caption{Multi-element}\label{fig:res:multi-element}
\end{subfigure}
  \begin{subfigure}[b]{0.49\linewidth}
  \centering
  \footnotesize
  \begin{gnuplot}[terminal=epslatex, terminaloptions={size 2.7,2.7 font 8}]
      set key right #above box opaque
      set key spacing 1.5
      # set key width -1

      #set style data histograms
      #set style histogram rowstacked

      #set boxwidth 0.5
      #set style fill solid

      set ylabel "Percentage"
      set xlabel "Max. Elements per Point"
      set yrange [0:100]
      #set xrange [-0.05:3.65]
      
      #set xtic offset 2,0 #rotate by -45 scale 0

      plot 'data/tree-multicontext-dmin.dat' using 1:($2+$3+$4+$5) t 'Partially Context Correct' with filledcurves x1 lc '#f8d491',\
      '' using 1:($2+$3+$4) t 'Partially Element Correct' with filledcurves x1 lc '#F1AA24',\
      '' using 1:($2+$3) t 'Fully Context Correct' with filledcurves x1 lc '#004c19',\
      '' using 1:($2) t 'Fully Element Correct' with filledcurves x1 lc '#009933'
  \end{gnuplot}
  \caption{Multi-context}\label{fig:res:multi-context}
\end{subfigure}
\caption{PCT with multiple land usage elements associated with each trajectory point}\label{fig:res:multi}
\end{figure}

The PCT is capable of predicting multiple elements and contexts at the same time. This may be useful in instances where a user is within, for instance, a building that is contained within another building (e.g.\ a shop within a shopping centre). Evaluating predictions made by trees that allow multiple such elements to be associated with a single time is conducted in accordance with the metrics defined in Section~\ref{sect:meth:evaluation}. Multi-element prediction is shown in Figure~\ref{fig:res:multi-element} for different maximum numbers of elements per point. Increasing the number of elements decreases the ability for the PCT to identify exactly which elements are being interacted with, however, the \emph{partial} value remains fairly consistent, demonstrating that the PCT typically gets some of the predictions correct in times when it cannot predict all elements correctly. Figure~\ref{fig:res:multi-context} shows the same graph, but for multi-context prediction where \emph{fully element correct} indicates that the set of elements being interacted with was predicted correctly, \emph{fully context correct} represents times when a prediction covers all correct elements through a parent context. \emph{Partially element correct} indicates that some elements were correctly predicted, but either additional elements were included in the prediction or some elements were overlooked, and \emph{partially context correct} means that some contexts that were predicted were correct, but again, not all elements are covered by a context or additional contexts are predicted.

\section{Conclusion}\label{sec:conclusion}

This work has explored the potential for replacing extracted locations with land usage elements that represent real-world features for prediction applications. After demonstrating the improved accuracies that this technique affords, the work has also extended the Context Tree data structure to create the Predictive Context Tree, a hierarchical classification model that is capable of both predicting future land usage interactions with comparable accuracy to existing techniques, but also predicting future contexts that users will be immersed in. Focusing on context prediction in a manner that requires only geospatial trajectories to be collected from the users provides the basis for a wealth of smart applications and services without the invasive data collection typically required by existing context identification work.

Through an evaluation over real-world data from two datasets, and a comparison to existing approaches, we have demonstrated the utility afforded by the PCT. Furthermore, properties of the predictions and the impact of the parameters have been explored. The primary contributions of this paper have been: (i) the proposal and evaluation of replacing extracted locations with identified land usage elements through a modified identification procedure, (ii) the PCT classification model, (iii) an evaluation of the PCT over real-world data demonstrating its ability to predict both interactions and contexts, and (iv) a parameter exploration to understand how properties of the PCT impact predictions.

This work has demonstrated many advantages of predicting over identified land usage elements, in addition to the advantages offered by the PCT. Future work, however, should explore the possibility of using techniques other than SVMs to further increase the predictive accuracy of the PCT, aiming to achieve even higher accuracy for element and context correct predictions.


\section*{Acknowledgements}

Portions of the research in this paper used the MDC Database made available by Idiap Research Institute, Switzerland and owned by Nokia~\cite{Kiukkonen:2010vm,Laurila:2012vk}, in addition to data from members of the Department of Computer Science, University of Warwick.

\bibliographystyle{WarwickBibliography}
\bibliography{bibliography} 

\end{document}